%% file: main.tex
\documentclass[10pt,twocolumn,letterpaper]{article}

\usepackage{cvpr}
\usepackage{times}
\usepackage{epsfig}
\usepackage{graphicx}
\usepackage{amsmath}
\usepackage{amssymb}

\usepackage{multirow} 
\usepackage{multicol} 
\usepackage{arydshln}
\usepackage[table]{xcolor}
\usepackage{subfloat}
\usepackage{subfigure}
\usepackage{rotating}
\usepackage{comment}
\usepackage{arydshln}
\usepackage{makecell}
\usepackage{authblk}
\usepackage{url}
\usepackage{booktabs}

\let\svthefootnote\thefootnote
\newcommand\blankfootnote[1]{%
  \let\thefootnote\relax\footnotetext{#1}%
  \let\thefootnote\svthefootnote%
}

\newcommand{\lidar}{\mbox{LiDAR }}

\usepackage[pagebackref=true,breaklinks=true,letterpaper=true,colorlinks,bookmarks=false]{hyperref}

\cvprfinalcopy % *** Uncomment this line for the final submission

\def\cvprPaperID{9150} % *** Enter the CVPR Paper ID here

\ifcvprfinal\fi
\begin{document}

%%%%%%%%% TITLE
%\title{PolarNet: An Improved Grid Representation of \lidar Point Clouds \\ for Semantic Segmentation}
\title{PolarNet: An Improved Grid Representation for \\ Online LiDAR Point Clouds Semantic Segmentation}

\author[1]{\textbf{Yang~Zhang}\thanks{Contributed equally.}}
\newcommand\CoAuthorMark{\footnotemark[\arabic{footnote}]} % get the current value
\author[1]{\textbf{Zixiang~Zhou}\protect\CoAuthorMark}
\author[2]{\textbf{Philip~David}}
\author[3]{\textbf{Xiangyu~Yue}}
\author[1]{\textbf{Zerong~Xi}}
\author[1]{\textbf{Boqing~Gong}\thanks{Now at Google.}}
\author[1]{\textbf{Hassan~Foroosh}}
\affil[1]{
\small{Department of Computer Science, University of Central Florida}
}
\affil[2]{
Computational and Information Sciences Directorate, U.S. Army Research Laboratory
}
\affil[3]{
Department of Electrical Engineering and Computer Sciences, University of California, Berkeley
}
\affil[ ]{\tt yangzhang@knights.ucf.edu, zhouzixiang@knights.ucf.edu, philip.j.david4.civ@mail.mil, xyyue@berkeley.edu, zxi@knights.ucf.edu, boqinggo@outlook.com, foroosh@cs.ucf.edu}

\maketitle
\thispagestyle{empty}

\blankfootnote{Code at  \url{https://github.com/edwardzhou130/PolarSeg}}

%%%%%%%%% ABSTRACT
\begin{abstract}
The need for fine-grained perception in autonomous driving systems has resulted in recently increased research on online semantic segmentation of single-scan LiDAR. Despite the emerging datasets and  technological advancements, it remains challenging due to three reasons: (1) the need for near-real-time latency with limited hardware; (2) uneven or even long-tailed distribution of LiDAR points across space; and (3) an increasing number of extremely fine-grained semantic classes. In an attempt to jointly tackle all the aforementioned challenges, we propose a new \lidar \hspace{-4.2pt}-specific, nearest-neighbor-free segmentation algorithm --- \textbf{PolarNet}. Instead of using common spherical or bird's-eye-view projection, our polar bird's-eye-view representation balances the points across grid cells in a polar coordinate system, indirectly aligning a segmentation network's attention with the long-tailed  distribution of the points along the radial axis. We find that our encoding scheme greatly increases the mIoU in three drastically different segmentation datasets of real urban \lidar single scans while retaining near real-time throughput.
   %With the increasing needs of fine-grained perception in autonomous driving, lidar semantic segmentation starts to receives more attention. Recent emerging datasets and technological advancement enable researchers to benchmark this problem and improve the algorithm. In this paper, we specifically address the in-demand street-view lidar segmentation. Comparing with generic point cloud semantic segmentation, urban lidar segmentation 1, prefers near real-time latency with limited hardware; 2, processes points that distribute unevenly across space 3, predicts increased and more fine-grained classes. The combination of aforementioned challenges motivates us to propose a new lidar-specific semantic segmentation algorithm - ?. Instead of using common handcrafted features, our polar bird-eye-view representation encodes more information. We find such encoding scheme greatly increase the mIoU in real urban lidar segmentation dataset while retaining near real-time FPS.
\end{abstract}
%%%%%%%%% BODY TEXT

\input{intro}

\input{related}
\input{approach}
\input{exp}

\section{Conclusion}

In this paper, we present a novel data representation for the online, single-scan \lidar point cloud semantic segmentation problem. Our approach addresses the problem of long-tailed spatial distribution of \lidar point clouds by quantizing points into polar bird's-eye-view (BEV) grids, where we encode points into fixed size representations through a trainable PointNet. Built upon the polar grid representation, our PolarNet network achieves a significant improvement in mIoU over state-of-the-art methods on the SemanticKITTI, A2D2, and Paris-Lille-3D datasets with fewer parameters, more throughput, and lower inference latency. Moreover, our experiments show universal improvement among different segmentation networks using our polar BEV compared to spherical projection and Cartesian BEV, indicating that our polar grid is a superior yet general \lidar point cloud data representation for the online semantic segmentation problem.

\clearpage
{\small
\bibliographystyle{ieee_fullname}
\bibliography{cvpr2020}
}

\end{document}

%% file: intro.tex
%!Tex root = main.tex

\section{Introduction}
There has been a great surge of \lidar point cloud data over the last decade, especially in the self-driving domain. In order to make use of the \lidar point clouds in various downstream applications, it is vital to develop automatic analytic methods to make sense of the data. In this paper, we focus on the online fine-grained semantic segmentation of \lidar point clouds. Similar to image semantic segmentation, the task is to assign a semantic label to each of the points given an input point cloud. 

\begin{figure}
    \centering
    \includegraphics[width = 0.8\linewidth]{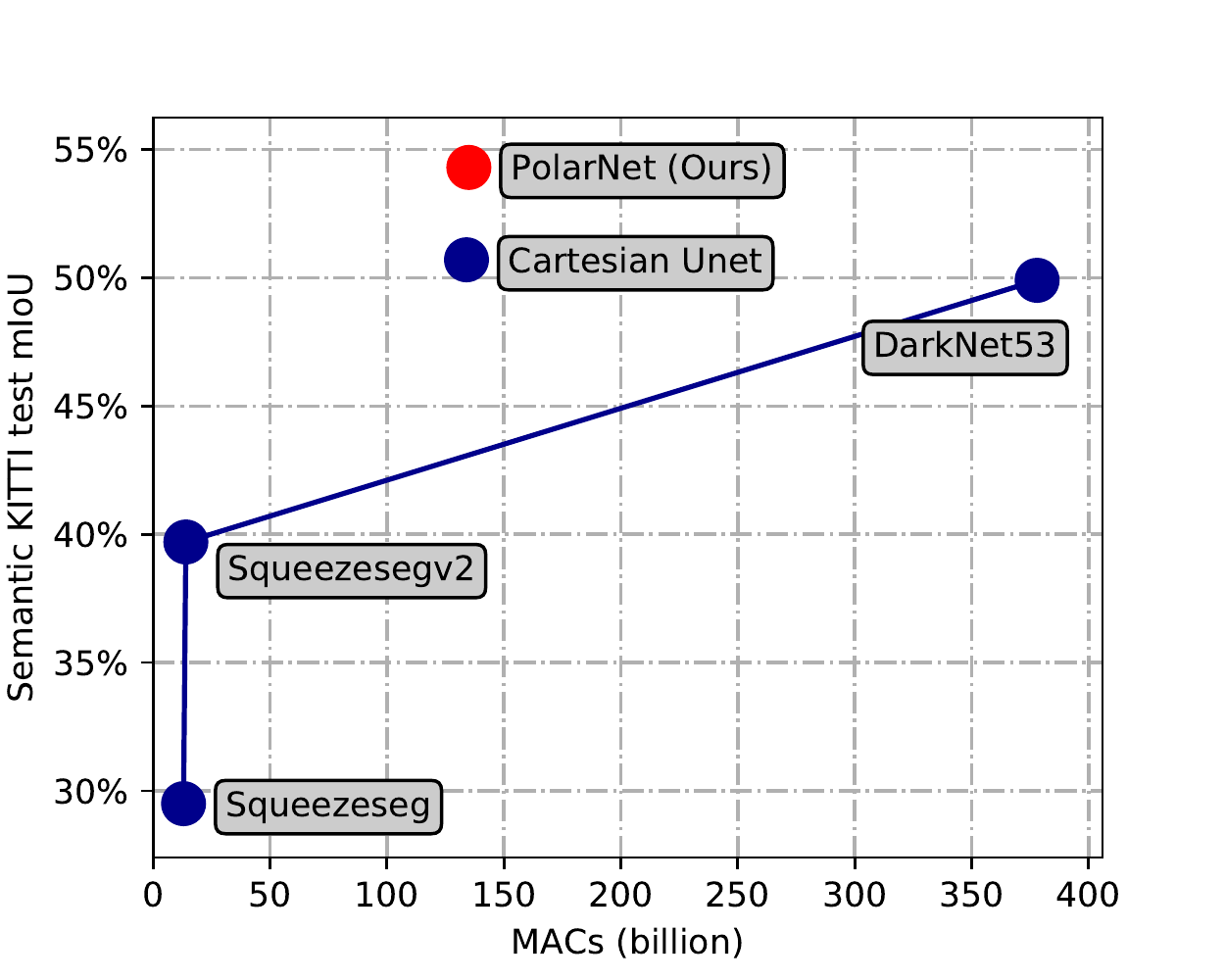}
    \caption{Point-level SemanticKITTI~\cite{behley2019iccv} segmentation mIoU vs.  multiply–accumulate operations per scan on the same GPU. Our Unet-based PolarNet not only significantly outperforms Cartesian-BEV Unet, PointNet, SqueezeSeg and SqueezeSeg's overparameterized variants (connected by line), but also retains remarkably low computational cost.}
    \label{fig:teaser}
    \vspace{-15pt}
\end{figure}

While several large-scale \lidar point clouds datasets are publicly available~\cite{geiger2012cvpr,waymo_open_dataset,zhang2015sensor,nuscenes2019}, it is until recent that the semantic segmentation labels, provided by~\cite{behley2019iccv,aev2019}, are able to match their scales. The  lag between the release of massive point clouds and the readiness of semantic segmentation labels indicates the challenge for human raters to provide point-wise labels and the demand for automatic and fast semantic segmentation solutions for  \lidar scans. 

We consider to use end-to-end deep neural networks for the single-scan semantic segmentation of \lidar point clouds. Before studying the network architecture or advanced training algorithms, however, we first focus on the input to the network. What constitutes a good input representation of one \lidar point cloud scan? We draw inspirations from several related domains to answer this question. 

In image segmentation, the perception field~\cite{YuKoltun2016} is one of the most principled considerations in designing high-performing CNN. It determines how much context a neural network can ``perceive'' before it classifies a pixel to a semantic class. In general, large perception fields improve performance. Techniques to enlarge the perception fields of  convolutional neural networks include dilated convolution~\cite{YuKoltun2016,chen2017rethinking}, feature pyramid~\cite{lin2017feature}, etc.

When it comes to the \lidar point clouds, we conjecture that not only the size but also the shape of the perception field matters. If we view a \lidar scan from a bird's-eye view, the points are organized in rings of various radii (cf.\ Figures~\ref{fig:overview} and  \ref{fig.bev}). As a result, the regular Cartesian coordinate would distribute the points into the grid cells in a non-uniform manner. Cells that are close to the sensor have to condense many points by each cell, blurring out fine-details of the points. In contrast, cells that are far away from the sensor each contain very sparse points, supplying limited cues for the neural network to label the points in such a cell. 

To this end, we propose to let the CNN perception fields track the special ring structure by partitioning a \lidar scan with a polar grid.  This simple change of the representation of the input to a neural network turns out to be very effective, boosting various semantic segmentation networks' performances by significant margins. 

Existing works on the \lidar scan understanding, however, fail to track the ring structure. Wu et al.~\cite{wu2019squeezesegv2} convert the point-cloud segmentation problem to a depth map segmentation problem by spherically projecting points onto an image. Zhang et al.~\cite{zhang2018efficient} handcraft a bird's-eye-view (BEV) representation of the point cloud and yet represent it by regular grids. Yang et al.~\cite{yang2018pixor} employ a similar BEV representation for object detection from the \lidar point clouds. 

On the one hand, the works above show it is promising to employ BEV representations of the \lidar scans in segmentation and detection. On the other hand, however, we contend they fail to fully take advantage of the structures revealed from BEV. We boost the vanilla BEV representations in two major ways. One is the polar grid to track the ring structures in the \lidar scans. The other is that we learn, instead of handcrafting, the local features per grid cell.

While polar coordination is no stranger to pre-DL computer vision~\cite{belongie2001shape}, it is rare in CNN given the images as well as feature matrices are essentially Cartesian. To fully integrate the polar BEV representation with a 2D CNN, we first redesign the BEV quantization scheme. Instead of quantizing points based on their Cartesian coordinates on the $XY$ plane, we now assign points according to their top-down polar coordinates as shown in Fig.~\ref{fig.bev}. Mimicking BEV's circular pattern with increasing sparsity, polar BEV significantly balance the points per grid by near one order of magnitude (c.f. Fig.~\ref{fig:pt_vs_dist}). Inspired by Lang et al.\cite{lang2019pointpillars}, we then learn a simplified PointNet to transform points in each grid into a fix-length representation vector. 

Since we quantize the points in polar coordinate, ideally the feature matrix should be in polar coordinate as well. To ensure the consistency of the perception field in the downstream CNN, we arrange those feature vectors into a polar grid whose leftmost and rightmost column are connected. We also modified the downstream CNN to be capable to convolve continuously on the polar grid. After obtaining the discrete prediction, which is also a polar grid, we map it back to the points in Cartesian space and evaluate the performance. Our pipeline is visualized in Fig.~\ref{fig:overview}.

We validate our approach on SemanticKITTI\cite{behley2019iccv}, A2D2\cite{aev2019} and Paris-Lille-3D\cite{roynard2018paris} datasets. Results show that our approach outperforms the state of art method by 2.1\%, 4.5\% and 3.7\%, respectively, on mean intersection-over-union (mIoU) evaluation metric with merely 1/3 of its parameters and MACs. 

The contributions of our work are summarised as follows:

\begin{itemize}
    \item We propose a more suitable \lidar scan representation which takes the imbalanced spatial distribution of points into consideration.
    \item Our presented PolarNet network, which is trained end-to-end using our polar grid data representation, surpasses the state of art method on public benchmarks withlow computational cost as shown in Fig.~\ref{fig:teaser}.
    \item We provide thorough analysis on the semantic segmentation performance based on different backbone segmentation networks using a polar grid compared to other representations, such as Cartesian BEV. 
    
\end{itemize}

%% file: related.tex
%!Tex root = main.tex

\section{Related Works}

\subsection{Point cloud applications and methods}
%\PD{What are general point clouds? How do they differ from Lidar point clouds?}
Most current point cloud applications focus on general point clouds in which points are densely distributed on object surfaces, such as single 3D object shape recognition~\cite{wang2019dynamic}, indoor point cloud segmentation~\cite{tchapmi2017segcloud,shi2019pointrcnn}, and reconstruction of outdoor scenes from point clouds~\cite{Tatarchenko_2018_CVPR}.
Despite sharing different tasks, in order to reach their goals, they must address a similar core problem: how to extract contextual information, whether local or global, from points that are irregularly distributed in space. Judging by the approach of aggregating context information, there are mainly two ways this is done: parameterized~\cite{wang2019dynamic,velickovic2018graph,landrieu2018large,huang2018recurrent} and non-parameterized~\cite{qi2017pointnet,qi2017pointnet++,shi2019pointrcnn}. Other works voxelize the points and then apply a 3D volume segmentation /detection algorithm~\cite{tchapmi2017segcloud}.
The representative work of the latter approach is the famous PointNet~\cite{qi2017pointnet} algorithm. PointNet and its successor~\cite{qi2017pointnet++} individually process each point and then use a set function to aggregate context information among those points. The parameterized ones are more commonly seen in the graph-based approaches~\cite{wang2019dynamic,velickovic2018graph,landrieu2018large}, where the points are modeled as a graph via KNN and then convoluted based on their graph connectivity.

\subsection{\lidar applications and methods}

%The resource is much more restricted in \lidar scenarios, where real-time performance is required on on-device hardware. And this is where researcher proposed bird-eye-view as a trade-off representation~\cite{?}. Different methods have their own way to create BEV representation, but the idea is always quantizing point into a top-down snapshot of the scene to a 2D neural network to avoid using expensive graph neural network or 3D operation. Being an empirically better representation than depth map~\cite{?}, BEV representation is widely used on lidar detection~\cite{??} and recently segmentation~\cite{??}.
Although \lidar sensors provide highly accurate distance measurement regardless of lighting conditions, the point clouds generated from \lidar are more sparse in space, which makes it more challenging to extract information from. Besides, processing resources in systems where \lidar sensors are typically used, such as in self-driving vehicles, are often restrictive, requiring real-time performance from embedded hardware. To address this issue, researchers have proposed different representations for the 3D data, which can be categorized into front-view and bird's-eye-view (BEV). Although different representations of the \lidar 3D point clouds are used, each quantizes the points into a compressed 2D snapshot of the scene that may be processed by a 2D neural network, thus avoiding expensive graph neural networks or 3D operations. 

Front view representations include depth image-like and spherical projections. Depth map or viewing frustum approaches apply a pinhole camera model to project 3D point clouds onto a 2D image grid. \cite{qi2018frustum} clustered points according to the frustum, where a 3D deep neural network is used within to identify the object. In spherical projection, points are projected onto a 2D spherical grid for a dense representation. SqueezeSeg~\cite{wu2018squeezeseg} and SqueezeSegV2~\cite{wu2019squeezesegv2} used spherical projections to represent point clouds for a light 2D semantic segmentation network, which is able to achieve real-time performance. The prediction result is further smoothed through a conditional random field (CRF) model and then re-projected back to a 3D point cloud. RangeNet++~\cite{milioto2019rangenet++} replaced the backbone network of SqueezeNet and CRF in SqueezeSeg to YOLOv3~\cite{redmon2018yolov3} Darknet and a GPU-based K-nearest neighbor search to achieve a better segmentation result. Being an empirically better representation than the depth map, BEV represents point clouds from a top-down perspective without losing any scale and range information and is widely used for \lidar detection~\cite{wang2019pseudo,yang2018pixor,lang2019pointpillars,ku2018joint,yan2018second} and recently also for segmentation~\cite{zhang2018efficient}. PIXOR~\cite{yang2018pixor} encoded the feature of each cell after discretizating point clouds into BEV representation as occupancy and normalized reflectance. Next, a neural network with 2D convolutional layers is used for 3D object detection. PointPillars~\cite{lang2019pointpillars} improved this idea by adding a PointNet model on the BEV representation.

There are many \lidar object detection datasets in existence, such as the Waymo Open Dataset~\cite{waymo_open_dataset} and the KITTI 3D detection dataset~\cite{geiger2012cvpr}. \lidar scan semantic segmentation datasets, conversely, are somewhat rare. To our knowledge, there are only three so far: the Audi dataset~\cite{aev2019}, Paris-Lille-3D~\cite{roynard2018paris} and the Semantic KITTI dataset~\cite{behley2019iccv}. Other point cloud segmentation datasets, such as Semantic3D~\cite{hackel2017semantic3d}, are out of the scope of online \lidar segmentation. Annotating RGB images for semantic segmentation algorithm development is a laborious task; however, the task of annotating \lidar data for semantic segmentation is even more difficult and less intuitive, which might be the reason for so few \lidar segmentation datasets.

\subsection{2D semantic segmentation}
2D semantic segmentation networks, which evolved from Fully Convolutional Networks (FCN)~\cite{long2015fully}, have demonstrated a significant improvement on various benchmarks in recent years. Similar to the success in other computer vision tasks, such as pose estimation and object detection, most efficient semantic segmentation networks~\cite{Yu2017} adopt an encoder-decoder structure, where a 2D image feature map is first reduced to extract high level contextual information and then expanded to retrieve spatial information. Among these networks, DeepLab~\cite{chen2014semantic} and Unet~\cite{ronneberger2015u} are two well-known successful representatives, both of which are designed to fuse multi-scale contextual information together. DeepLab and its successors~\cite{chen2017rethinking,chen2018encoder} took advantage of diluted convolution filters to increase the reception field while Unet added skip connections to directly concatenate different levels of semantic features and is proven to be more efficient in images with irregular and coarse edges, like medical images.

%% file: approach.tex
\section{Approach}

\begin{figure*}
  \centering
  \includegraphics[width=0.8\textwidth]{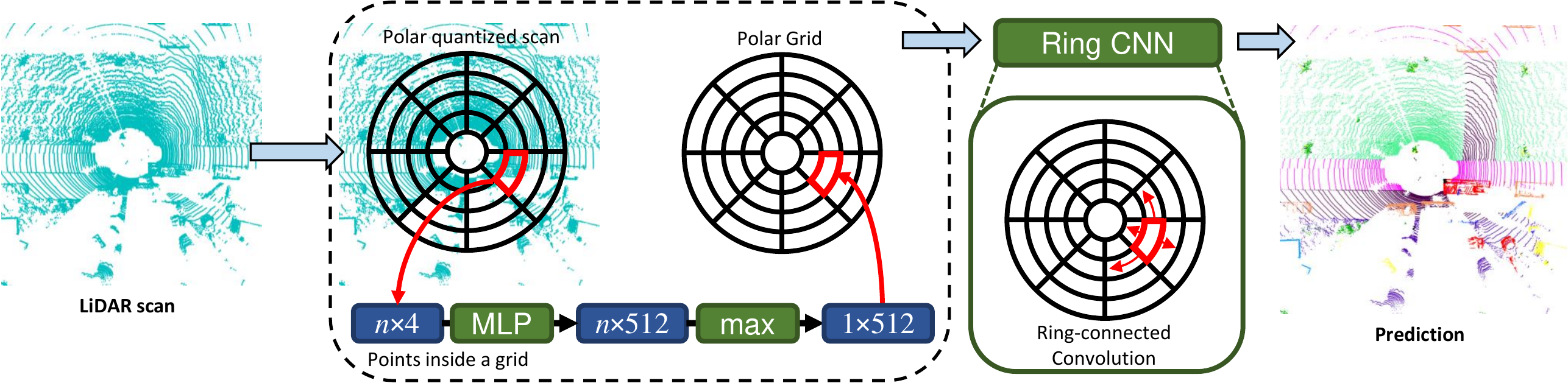}
  \caption{Overview of our model. For a given \lidar point cloud, we first quantize the points into grids using their polar BEV coordinates. For each of those grid cells, we use a simplified KNN-free PointNet to transform points in it to a fixed-length representation. The representation is then assigned to its corresponding location in the ring matrix. We input the matrix to the ring CNN, which is composed of ring convolution modules. Finally, the CNN outputs a quantized prediction and we decode it to the point domain.}
  \label{fig:overview}
  \vspace{-15pt}
\end{figure*}

\begin{figure}[htbp]
\centering
\subfigure[Cartesian BEV]{
\includegraphics[width=3.4cm]{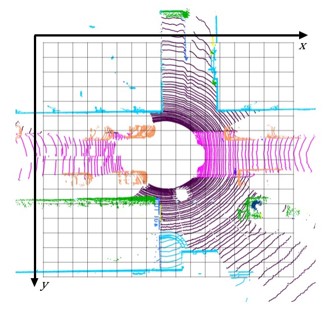}
\label{fig.t_bev}
}
\quad
\subfigure[Polar BEV]{
\includegraphics[width=3.4cm]{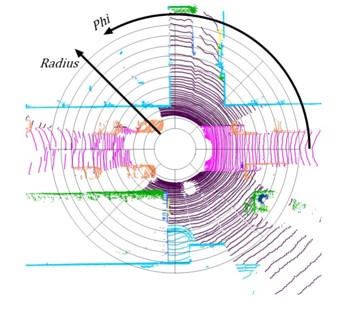}
}
\quad
\caption{Two BEV quantization strategies. Each grid cell on the image denotes one feature in a feature map.}
\vspace{-15pt}
\label{fig.bev}
\end{figure}

\subsection{Problem Statement}

Given a training dataset of $N$ \lidar scans $\{(P_i,L_i)|i=1,\ldots,N\}$, $P_i\in\mathbb{R}^{n_i\times4}$ is the $i$th point set containing $n_i$ \lidar points. Each row of $P_i$ consists of four features representing one \lidar point $p$, namely ($x$, $y$, $z$, reflection). ($x$, $y$, $z$) is the Cartesian coordinate of the point relative to the scanner. The reflection is the intensity of returning laser beam. $L_i\in\mathbb{Z}^{n_i}$ contains the object labels for each point $p_j$ in $P_i$.

Our goal is to learn a segmentation model $f(\cdot \ ;\theta)$ parameterized by $\theta$ so that the difference between the prediction $f(P_i)$ and $L_i$ is minimized.

\subsection{Bird's-eye-view Partitioning}

Although a point cloud scan consists of scattered observations of the surrounding 3D environment, empirically, one may represent it as a top-down snapshot of the scene with minimum information loss. \cite{chen2017multi} proposes to input such top-down orthogonal projections directly into a 2D detection network to detect objects in 3D point clouds. And it is later on used in point cloud segmentation~\cite{zhang2018efficient}. By taking a 2D top-down image as the input, the network outputs a tensor of the same dimensional shape with each spatial location encoding the class prediction for each voxel along the z-axis of that location. This elegant approach accelerates the segmentation process by taking advantage of years of research in 2D CNNs. It also avoids expensive 3D segmentation and 3D graph operations.

The original motivation of the BEV was to represent the scene with a top-down image to speed up the downstream task-specific CNNs. Based on years of experience designing CNN architectures, researchers choose BEV representations to closely resemble the appearance of natural images so as to maximally utilize the downstream CNNs, which happen to be designed for natural images. Hence, initial BEV representations created top-down projections of the point clouds. Recently, variants of the initial BEV attempt to encode each pixel in the BEV with rich different heights~\cite{yang2018pixor}, reflection~\cite{simon2018complex} and even learned representations~\cite{lang2019pointpillars}. However, one thing remained unchanged: the BEV methods used a Cartesian grid partition as shown in Fig.~\ref{fig.t_bev}.

A grid is the fundamental image representation, but it may not be the best representation for BEV. A BEV is a compromise between performance and precision. By observing a BEV image, we immediately notice that points densely concentrated in the middle grid cells and peripheral grid cells stay totally empty. Uneven partitioning does not only waste computational power, but also limits feature representiveness for the center grid cells. Besides, points with different labels might be assigned to a single cell. The minor points' predictions will be suppressed by the majority in the output since the final prediction is on voxel-level. 

\begin{figure}
  \centering
  \includegraphics[width=0.6\linewidth]{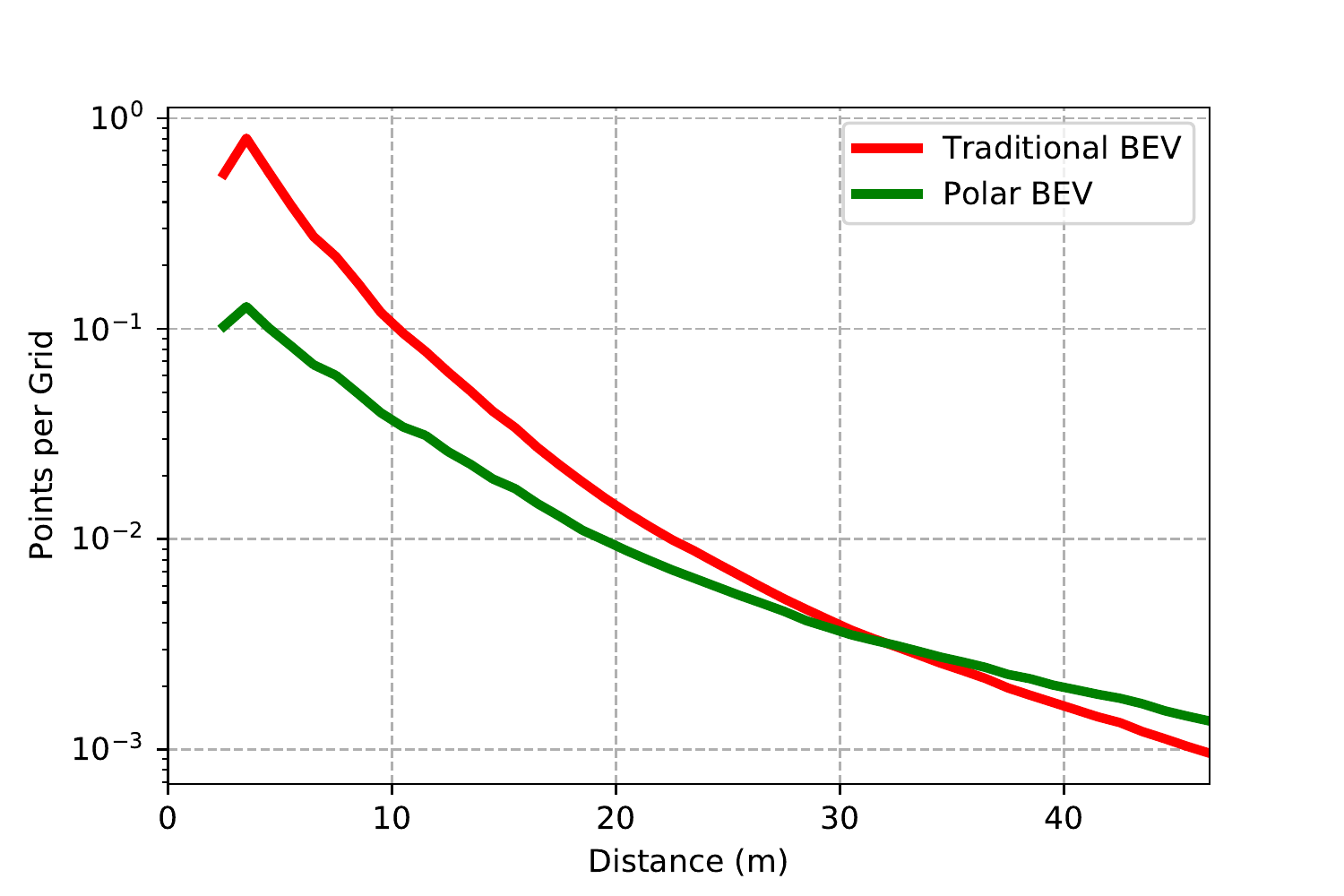}
  \vspace{-4pt}
  \caption{Grid cell distance from the sensor vs. logarithmically spaced mean number of points per grid cell. The traditional BEV representation allocates most of its grid cells to the further end with few points in them.}
  \label{fig:pt_vs_dist}
  \vspace{-15pt}
\end{figure}

\subsection{Polar Bird's-eye-view}

How do we address this imbalance? Based on the ring-like structure presented in the \lidar scan top-down view, we present our Polar partitioning replacing the Cartesian partitioning in Fig.~\ref{fig.bev}.

 Instead of quantizing points in a Cartesian coordinate system, we first calculate each point's azimuth and radius on the $XY$ plane with the sensor's location as the origin. We then assign points to grid cells based on their quantized azimuth and radius.
 
 We find the benefit of polar BEV to be twofold. First, it more evenly distributes the points. To verify this claim, we computed a statistic on the validation split of the SemanticKITTI dataset~\cite{behley2019iccv}. As shown in Fig.~\ref{fig:pt_vs_dist}, the points per polar grid cell is much less than in the Cartesian BEV when the cell is close to the sensor. This indicates that the representation for the densely occupied grid is finer.  With the same number of grid cells, the traditional BEV grid cell has on average $0.7\pm3.2$ points while polar BEV grid cell has on average $0.7\pm1.4$ points. The difference between the standard deviations indicates that, overall, the points are more evenly distributed across the polar BEV grids.
 
The second benefit of the polar BEV is that the more balanced point distribution lessens the burden on predictors. Since we reshape 2D network output to voxel prediction for point prediction, unavoidably, some points with different groundtruth labels will be assigned to the same voxel. And some of them will be misclassified no matter what. With the Cartesian BEV, on average, 98.75\% of points in every grid cell share the same label. And this number jumps to 99.3\% in the polar BEV. This indicates that points in the polar BEV are less subjected to misclassification due to the spatial representation. Considering that small objects are more likely to be overwhelmed by majority labels in a voxel, this 0.6\% difference might have a more profound impact in the eventual mIoU. To further investigate the mIoU upper bound, we set each point's prediction as the majority label of its assigned voxel. It turns out that the Cartesian BEV's mIoU reaches 97.3\% in the sanity check. And the polar BEV reaches 98.5\%. The higher upper bound in the polar BEV will likely increase the downstream model performance.
 
 \subsection{Learning the Polar Grid}

Instead of arbitrarily handcrafting the features for each grid, we capture the distribution of points in each grid with a fixed-length representation. It is produced by a learnable simplified PointNet~\cite{qi2017pointnet} $h$ followed by a max-pooling. The network only contains fully-connected layers, batch-normalization and ReLu layers. The feature in the $i,j$-th grid cell in a scan is:

\begin{equation}
    \textrm{fea}_{i,j} = \textrm{MAX}(\{h(p)| w_i<p_x<w_{i+1}, l_j<p_y<l_{j+1}\})
\end{equation}

\noindent where $w$ and $l$ are the quantization sizes. $p_x$ and $p_y$ are locations of point $p$ in the map. Note that the locations and quantization sizes could be either polar or Cartesian. We do not quantize the input point cloud along the z-axis. Similar to ~\cite{lang2019pointpillars}, our learned representation represents the entire vertical column of a grid.

If the representation is learned in the polar coordinate system, the two sides of the feature matrix will be connected along the azimuth-axis in physical space as shown in Fig.~\ref{fig:overview}. We developed a discrete convolution which we refer to as a ring convolution. The ring convolution kernel will convolve the matrix assuming the matrix is connected on both ends of the radius axis. Meanwhile, gradients located in the opposite side can propagate back to the other side through this ring convolution kernel. By replacing the normal convolution with the ring convolution in a 2D network, the network will be able to end-to-end process the polar grid without ignoring its connectivity. This provides models with extended receptive fields. Since it is a 2D neural network, the eventual prediction will also be a polar grid whose feature dimension equals to the multiplication of quantized height channel and number of classes. We can then reshape the prediction to a 4D matrix to derive a voxel-based segmentation loss.

As readers may notice, most CNNs are technically capable of processing polar grids if convolutions are replaced with ring convolutions. We refer to a network with ring-convolutions that is trained to process polar grids as a ring CNN.

%% file: exp.tex
%!Tex root = main.tex

\section{Experiments}

We present our experimental setup, results and ablation study in this section.

\subsection{Datasets}

We use the SemanticKITTI\cite{behley2019iccv}, A2D2\cite{aev2019} and Paris-Lille-3D\cite{roynard2018paris} datasets in our experiments.

\begin{figure}
\centering

\subfigure[SemanticKITTI]{
\includegraphics[width=0.26\linewidth]{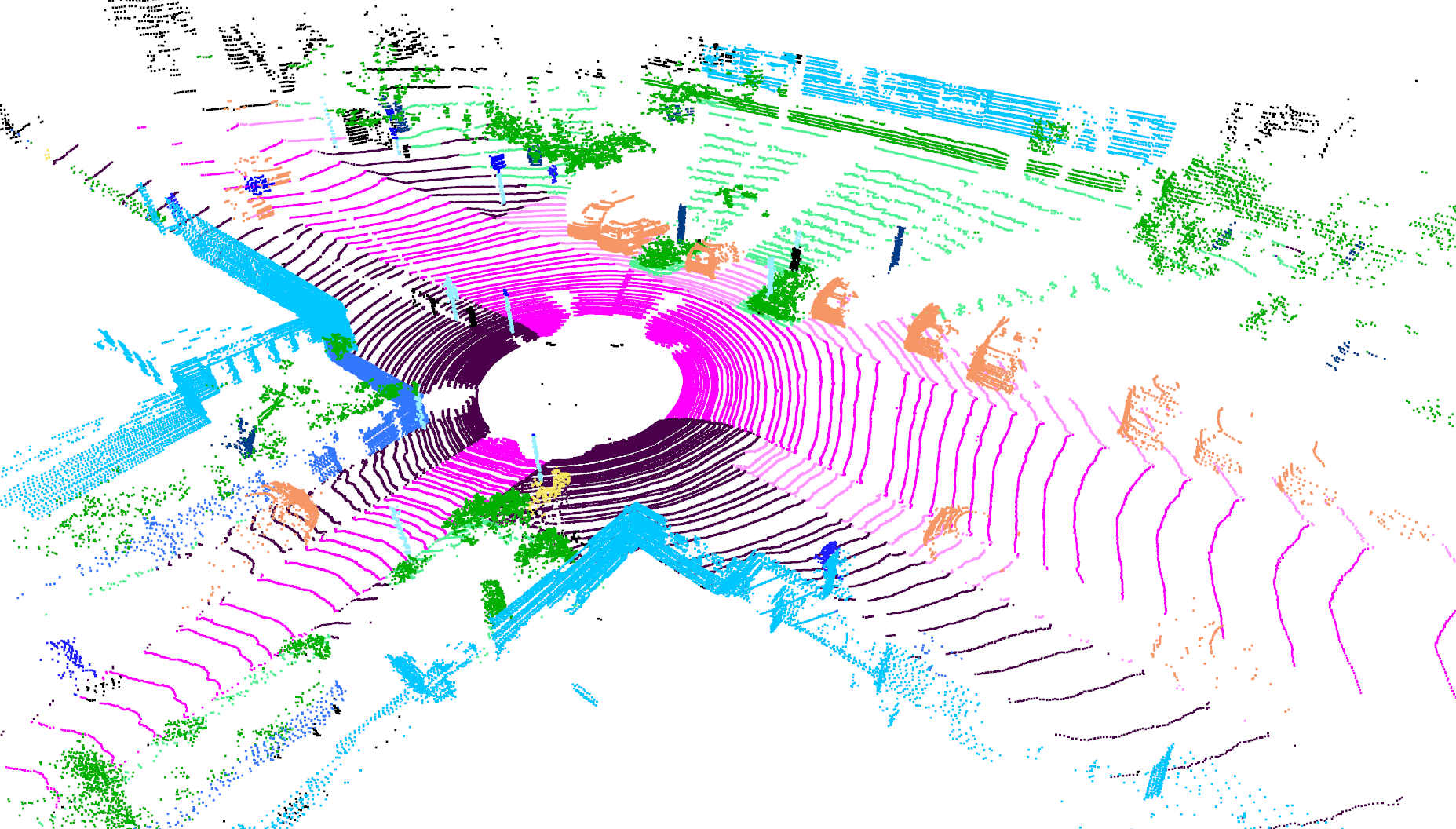}
 \label{fig:SKITTI}
}
\quad
\subfigure[A2D2]{
\includegraphics[width=0.26\linewidth]{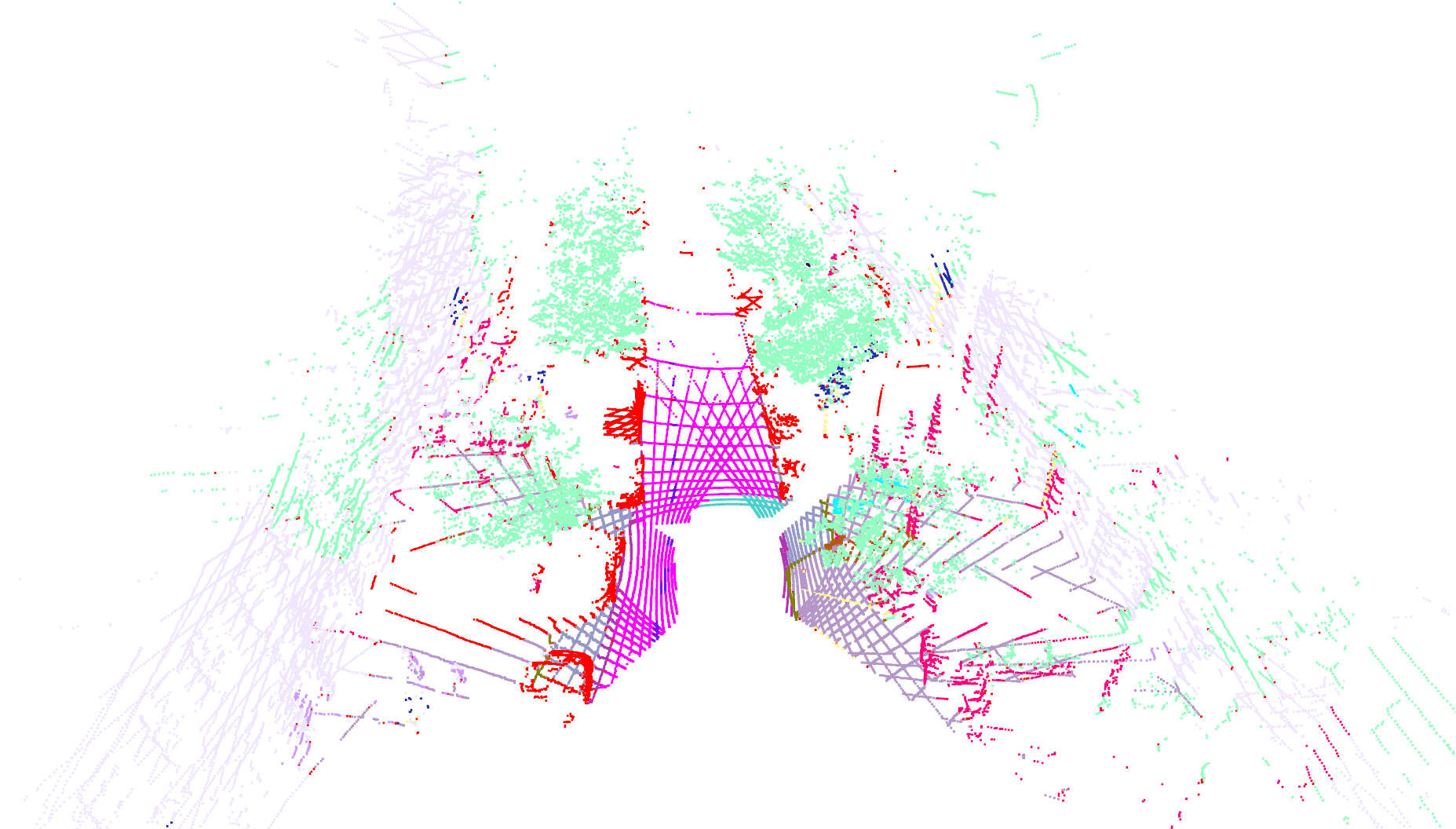}
 \label{fig:A2D2}
}
\quad
\subfigure[Paris-Lille-3D]{
\includegraphics[width=0.26\linewidth]{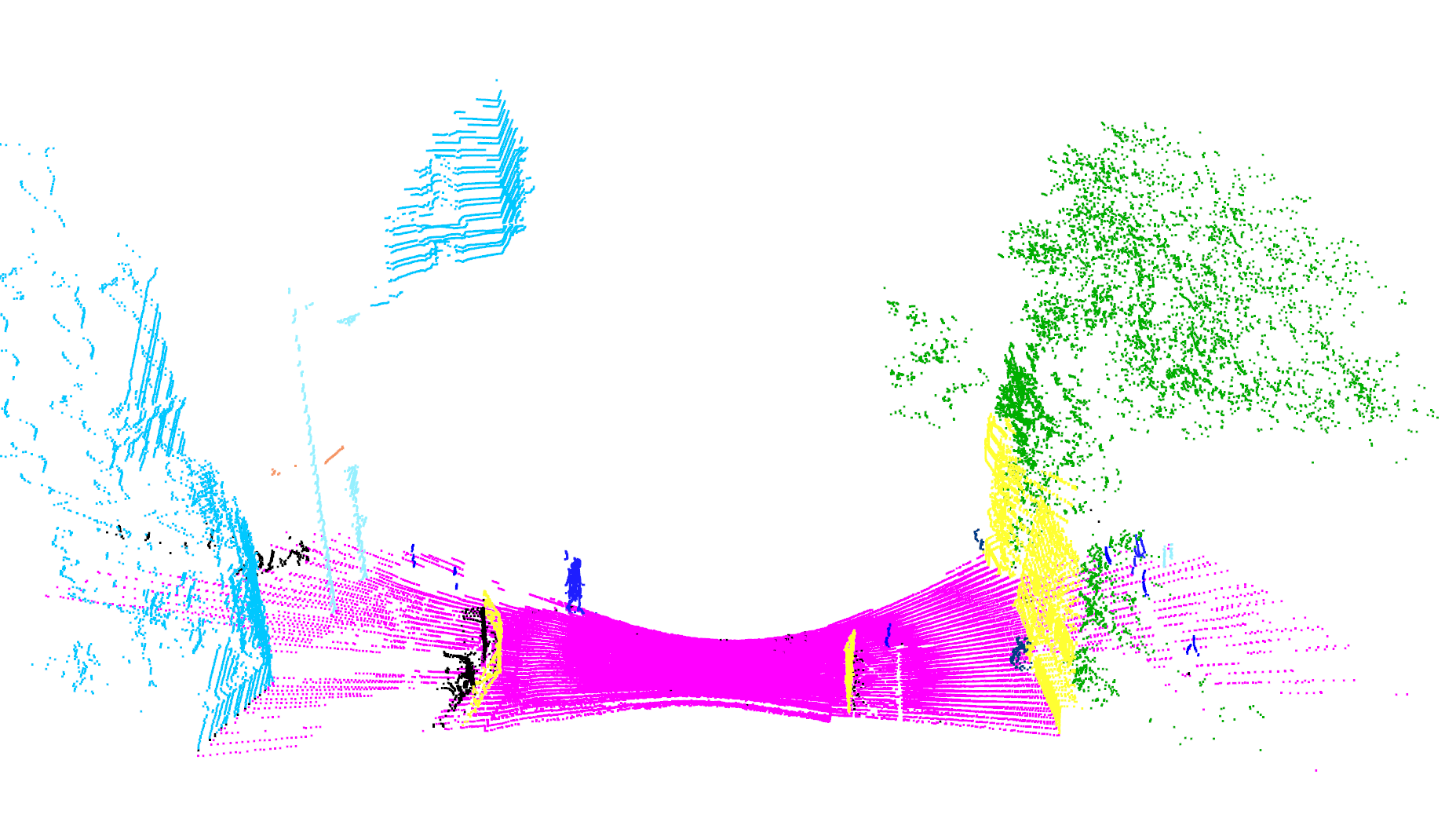}
\label{fig:PL}
}
\quad
\caption{PolarNet outperforms baselines despite different scanline patterns in datasets. Zoom in for more details.}
\vspace{-15pt}
\label{fig:dataset_visualization}
\end{figure}

\textbf{SemanticKITTI} is a point-level re-annotation of the \lidar part of the famous KITTI dataset~\cite{geiger2012cvpr}. It has a total of 43551 scans sampled from 22 sequences collected in different cities in Germany. It has 104452 points per scan on average and each scan is collected by a single  Velodyne HDL-64E laser scanner shown in Fig.\ref{fig:SKITTI}. There are 19 challenging classes in total. The most frequent class, `vegetation', has $4.82\times10^{7}$ times more points than the least frequency class, `motorcyclist'. Obviously, this is a heavily imbalanced and challenging dataset. We follow SemanticKITTI’s subset split protocol and use ten sequences for training, one for validation and the rest of them for testing. We present several baselines that have been presented with SemanticKITTI.  We report the segmentation performance on the SemanticKITTI testing subset by uploading our segmentation prediction to their evaluation server.

\textbf{A2D2} dataset is a comprehensive autonomous driving dataset developed by Audi. It includes  a 38-class segmentation annotation. Despite that the A2D2 data is presented as 3D points in space, these points distribute differently from the KITTI counterparts. We present an example in Fig.~\ref{fig:A2D2}. First of all, a single  sensor creates a  panoramic \lidar scan in the KITTI dataset. Meanwhile, A2D2 uses five asynchronous \lidar sensors where each sensor covers a potion of the surrounding view. Hence almost all the A2D2 reconstructed \lidar views do not cover all degrees. Secondly, as shown in Fig.~\ref{fig:A2D2}, A2D2 \lidar sensors do not necessarily produce horizontal scan-lines. Our goal is to simulate a vehicle's immediate perception during operation. We first project all \lidar points back to the vehicle coordinate system. We then manually create (semi-)panoramic \lidar compositions from any partial scans asynchronously generated within a time window of 50ms. Since sensors are not available all of the time, some generated scans are left incomplete. This heterogeneous composition poses a great challenge for all segmentation algorithms, including ours.  With the aforementioned \lidar panoramic stitching, we create 22408, 2774 and 13264 training, validation and test scans, respectively.

In contrast to the other two datasets, \textbf{Paris-Lille-3D} provides 3 aggregated point clouds, which are built from continuous \lidar scans of streets in Paris and Lille collected with one tilted rear-mounted Velodyne HDL-32E. Each point is annotated with one of nine segmentation classes, its timestamp and its world coordinate. Given scanner trajectory and points' timestamps, we extract individual scans from the registered point clouds. We record one scan every 50ms. Each scan is made of points within +/- 100ms, e.g.~\ref{fig:PL}. In total, we create 5112, 1205 and 1273 training, validation and test scans, respectively. We upload the testing predictions for Paris-Lille-3D to their evaluation server to obtain the official testing results. Since Paris-Lille-3D accepts composition predictions only, we aggregate multi-scan predictions via max-voting.

\textbf{Voxelization}: After analyzing the spatial distribution of points in the SemanticKITTI, A2D2 and Paris-Lille-3D training split, we respectively fixed the Cartesian BEV grid spaces to be $[x: \pm 50m, y: \pm 50m, z:-3 \sim 1.5m]$, $[x: \pm 50m, y: \pm 50m, z:-3 \sim 9m]$ and $[x: \pm 15m, y: \pm 15m, z:-3 \sim 12m]$ and respectively $[distance: 3 \sim 50m, z:-3 \sim 1.5m]$, $[distance: 0 \sim 50m, z:-3 \sim 9m]$ and $[distance: 0 \sim 15m, z:-3 \sim 12m]$ for our polar BEV to include more than 99\% of points for each scan on average. Points exceeding this range are assigned to the closest BEV grid cell. In addition, we set the respective grid sizes as $[480,360,32]$, $[320,320,32]$ and $[320,320,32]$.

\subsection{Baselines and Metric}

\textbf{SqueezeSeg}: As the pioneer work in this field, Wu et al.~\cite{wu2018squeezeseg} converted this problem to a 2D segmentation problem by projecting \lidar points onto a spherical surface surrounding the sensor. They also added a CRF to further improve the end results by enforcing the neighboring label consistency . Besides the vanilla SqueezeSeg and SqueezeSeg-v2, Behley et al.~\cite{behley2019iccv} replaced the SqueezeNet backbone with YOLO~\cite{redmon2018yolov3} Darknet-53. This over-parameterization further improved the results by more than 10\% on SemanticKITTI over SqueezeSeg-v2. In addition, RangeNet++~\cite{milioto2019rangenet++} includes a KNN-based post-processing method which is used after the CNN segmentation network to reduce the error created by the discretization of spherical intermediate representation.

\textbf{PointNet}\cite{qi2017pointnet}: PointNet is a simplistic network able to predict point semantic segmentation. It individually processes each point with a fully connected network first. Then it summaries a global representation by max pooling the features of all points. The predictor predicts each point’s class from the concatenation of that point’s features and the global representation. PointNet++~\cite{qi2017pointnet++} is an empirical improvement obtained by adding hierarchical pooling and context representation to vanilla PointNet.

\textbf{TangentConv}~\cite{Tatarchenko_2018_CVPR}: Tatarchenko and Park et al. propose to use tangent convolutions on surface geometry to predict segmentation classes for 3D point clouds.

\textbf{RandLA}~\cite{hu2019randla}: Hu et al. propose to segment large scale point clouds with a local feature aggregation module.

We report accuracy, per-class IoU and mIoU. mIoU is the mean over all semantic classes of class intersection over union. A class $c$'s intersection over union, $(IoU_c)$, refers to the intersection of the class prediction and ground truth divided by their union:

\begin{equation}
    IoU_{c} = \frac{|\mathcal{P}_{c}\cap \mathcal{G}_{c}|}{|\mathcal{P}_{c}\cup \mathcal{G}_{c}|}.
\end{equation}

Given the unique properties of \lidar applications, we also report models' single scan prediction latency, maximum frames-per-second with largest possible batch size (FPS), average multiply-accumulate operations per scan (MAC), and number of model parameters. We report the average on the entire validation split with the same GPU. We do not down-sample points in points-related models.

We use official implementations or reported results for our baselines. We implemented our own network in Pytorch~\cite{paszke2017automatic}. We use torch Geometric~\cite{Fey/Lenssen/2019} to parallelize points max pooling in each grid.

\begin{table*}
	\centering
	\caption{Segmentation results on \textbf{test} split of SemanticKITTI.}
	\label{tab:SKITTI_test}
	\resizebox{\linewidth}{!}{%
	\begin{tabular}{|l|c|c|c|c|c|c|*{19}{c}|}
		\hline
		\multirow{2}*{Model} & \multirow{2}*{FPS} & \multirow{2}*{Latency} & \multirow{2}*{MACs} & \multirow{2}*{Params} & \multirow{2}*{Acc} & \multirow{2}*{mIoU} & \multicolumn{19}{c|}{Per class IoU}  \\ 
		\cline{8-26}
		&&&&&&&\rotatebox{90}{car} &	\rotatebox{90}{bicycle} &	\rotatebox{90}{motorcycle} &	\rotatebox{90}{truck}&	\rotatebox{90}{other-vehicle} &	\rotatebox{90}{person} &	\rotatebox{90}{bicyclist} &	\rotatebox{90}{motorcyclist} &	\rotatebox{90}{road} &	\rotatebox{90}{parking} & \rotatebox{90}{sidewalk} &	\rotatebox{90}{other-ground} &	\rotatebox{90}{building} &	\rotatebox{90}{fence} &	\rotatebox{90}{vegetation} &	\rotatebox{90}{trunk} &	\rotatebox{90}{terrain} &	\rotatebox{90}{pole} &	\rotatebox{90}{traffic-sign} \\
		\hline 
		\hline
		
		PointNet~\cite{qi2017pointnet}& 11.5 & 0.087s & 141B & 3.5M & - & 14.6\% & 46.3\% &  1.3\% &  0.3\% & 0.1\% & 0.8\% & 0.2\% & 0.2\% & 0.0\% &  61.6\% & 15.8\% & 35.7\% &  1.4\% &  41.4\% & 12.9\% & 31.0\% & 4.6\% & 17.6\% &  2.4\% &  3.7\% \\
		
		PointNet++~\cite{qi2017pointnet++}& - & - & - & 6M & - & 20.1\% & 53.7\% &  1.9\% &  0.2\% &  0.9\% & 0.2\% & 0.9\% & 1.0\% & 0.0\% & 72.0\% &  18.7\% & 41.8\% &  5.6\% &  62.3\% &  16.9\% & 46.5\% &  13.8\% & 30.0\% &  6.0\% & 8.9\%\\
		
		Squeezeseg~\cite{wu2018squeezeseg}& 49.2 & 0.031s & 13B & 0.9M & - & 29.5\% & 68.8\% & 16.0\% & 4.1\% & 3.3\% & 3.6\% & 12.9\% & 13.1\% & 0.9\% & 85.4\% & 26.9\% & 54.3\% & 4.5\% & 57.4\% & 29.0\% & 60.0\% & 24.3\% & 53.7\% & 17.5\% & 24.5\%\\
		
		TangentConv~\cite{Tatarchenko_2018_CVPR}& - & - & - & 0.4M & - & 35.9\% & 86.8\% &  1.3\% &  12.7\% & 11.6\% & 10.2\% &  17.1\% &  20.2\% &  0.5\% &  82.9\% & 15.2\% &  61.7\% &  9.0\% &  82.8\% & 44.2\% &  75.5\% &  42.5\% &  55.5\% & 30.2\% & 22.2\%\\
		
		Squeezesegv2~\cite{wu2019squeezesegv2}& 36.7 & 0.036s & 14B & 0.9M & - & 39.7\% & 81.8\% & 18.5\% & 17.9\% & 13.4\% & 14.0\% & 20.1\% & 25.1\% & 3.9\% & 88.6\% & 45.8\% & 67.6\% & 17.7\% & 73.7\% & 41.1\% & 71.8\% & 35.8\% & 60.2\% & 20.2\% & 36.3\%\\
		
		DarkNet53~\cite{behley2019iccv}& 12.7 & 0.087s & 378B & 50M & 87.8\% & 49.9\% & 86.4\% & 24.5\% & 32.7\% & 25.5\% & 22.6\% & 36.2\% & 33.6\% & 4.7\% & \textbf{91.8\%} & 64.8\% & 74.6\% & \textbf{27.9\%} & 84.1\% & 55.0\% & 78.3\% & 50.1\% & 64.0\% & 38.9\% & 52.2\%\\
		
		RangeNet++~\cite{milioto2019rangenet++}& - & - & 378B & 50M & 89.0\% & 52.2\% & 91.4\% & 25.7\% & \textbf{34.4\%} &  25.7\% &  23.0\% & 38.3\% & 38.8\% & 4.8\% &  \textbf{91.8\%} &  \textbf{65.0\%} & \textbf{75.2\%} & 27.8\% & 87.4\% & 58.6\% &  80.5\% & 55.1\% & 64.6\% & 47.9\% &  55.9\%\\
		
		RandLA~\cite{hu2019randla}& - & - & - & 1.2M & -  & 53.9\% & \textbf{94.2\%} &  26.0\% &  25.8\% & \textbf{40.1\%} & \textbf{38.9\%} & \textbf{49.2\%} & \textbf{48.2\%} & \textbf{7.2\%}  & 90.7\% &  60.3\% & 73.7\% &  20.4\% & 86.9\% &  56.3\% &  81.4\% &  \textbf{66.8\%} & 49.2\% & 47.7\% & 38.1\%\\
		
		\hline
		
		Unet w/ Cartesian BEV& 19.7 & 0.051s & 134B & 14M & 87.6\% & 50.7\% & 92.7\% & 26.8\% & 23.1\% & 26.7\% & 24.2\% & 48.1\% & 41.0\% & 4.4\% & 86.7\% & 52.3\% & 67.2\% & 12.9\% & 89.5\% & 57.7\% & 80.8\% & 62.5\% & 62.5\% & 50.3\% & 53.5\%\\
		
		PolarNet& 16.2 & 0.062s & 135B & 14M & 90.0\% & \textbf{54.3\%} & 93.8\% & \textbf{40.3\%} & 30.1\% & 22.9\% & 28.5\% & 43.2\% & 40.2\% & 5.6\% & 90.8\% & 61.7\% & 74.4\% & 21.7\% & \textbf{90.0\%} & \textbf{61.3\%} & \textbf{84.0\%} & 65.5\% & \textbf{67.8\%} & \textbf{51.8\%} & \textbf{57.5\%}\\
		
		\hline
	\end{tabular}
	}
\vspace{-15pt}
\end{table*}

\subsection{SemanticKITTI Segmentation Experiment}

 Table \ref{tab:SKITTI_test} shows the performance comparison between our approaches and multiple baselines. The results demonstrate that our polar bird's-eye-view segmentation network based on Unet outperforms the state of the art method even with a smaller number of parameters and lower latency. As shown in this table, point-based methods like PointNet and TangentConv are inefficient when used with large \lidar point clouds and poor in segmentation accuracy. For per class IoU, our BEV approaches achieves improvements in most classes, especially in those classes that are irregular and sparsely distributed in space, which matches with the scale and range preserving properties of the polar BEV. We also notice particularly low performance on ``other-ground'' and ``motorcyclist.'' Investigation suggests they are visually indistinguishable from other classes. By SemanticKITTI's definition, ``other-ground'' is essentially sidewalk/terrain like ground but serving other purposes, e.g.,  traffic islands. As for motorcyclist, it is challenging even for a human to tell a motorcyclist from person or bicyclist because the motorcycle itself is often largely occluded. Besides, motorcyclists are the rarest class in the dataset --- constitute 0.004\% of the training points and only one instance appears in the official validation sequence.

\subsection{A2D2 Segmentation Experiment}

\iffalse
\begin{figure}[htbp]
\centering
\subfigure[SemanticKITTI]{
\includegraphics[width=0.75\linewidth]{fig/skitti_vis_2.png}
\label{fig.skitti_vis}
}
\quad
\subfigure[A2D2]{
\includegraphics[width=0.75\linewidth]{fig/a2d2_vis_2.png}
}
\quad
\caption{Point cloud visualization of SemanticKITTI and A2D2 datasets. One can clearly see how A2D2's 
non-vertical \lidar scanlines are entangled.}
\label{fig.a2d2_vis}
\end{figure}
\fi

\begin{table*}[htp]  
	\centering
	\caption{Segmentation results on \textbf{test} split of A2D2.}
	\label{tab:A2D2_test}
	\resizebox{\linewidth}{!}{%
	\begin{tabular}{|l|c|c|c|c|c|c|*{17}{c}|}
		\hline
		\multirow{2}*{{Model}} & \multirow{2}*{{FPS}} & \multirow{2}*{{Latency}} & \multirow{2}*{{MACs}} & \multirow{2}*{{Params}} & \multirow{2}*{{Acc}} & \multirow{2}*{{mIoU}} & \multicolumn{17}{c|}{Per class IoU} \\ 
		\cline{8-24}
		&&&&&&& \rotatebox{90}{car} & \rotatebox{90}{bicycle} & \rotatebox{90}{pedestrian} &	\rotatebox{90}{truck}&
		\rotatebox{90}{\makecell{small \\ vehicles} } &
		\rotatebox{90}{\makecell{traffic \\ signal}} &
		\rotatebox{90}{\makecell{traffic \\ sign}} &	\rotatebox{90}{\makecell{utility \\ vehicle}} &
		\rotatebox{90}{sidebars} &	\rotatebox{90}{\makecell{speed \\ bumper}} & \rotatebox{90}{curbstone} &	\rotatebox{90}{solid line} &	\rotatebox{90}{\makecell{irrelevant \\ signs}} &
		\rotatebox{90}{\makecell{road \\ blocks}} &
		\rotatebox{90}{tractor} &
		\rotatebox{90}{\makecell{non-\\drivable \\ street} } &
		\rotatebox{90}{\makecell{zebra \\ crossing}}\\
		\hline 
		\hline
		
		Squeezeseg~\cite{wu2018squeezeseg}& 87.5 & 0.009s & 15B & 0.9M & - & 8.9\% & 9.7\% & 0.0\% & 0.0\% & 15.8\% & 0.0\% & 0.7\% & 64.4\% & 0.0\% & 0.4\% & 0.0\% & 2.2\% & 15.6\% & 0.5\% & 15.9\% & 0.0\% & 0.0\% & 0.0\% \\

		Squeezesegv2~\cite{wu2019squeezesegv2}& 67.1 & 0.015s & 15B & 0.9M & 81.0\% & 16.4\% & 15.4\% & 0.2\% & 8.6\% & 63.8\% & 0.0\% & 16.8\% & 61.7\% & 0.6\% & 0.1\% & 0.0\% & 14.8\% & 24.7\% & 12.7\% & 33.2\% & 0.0\% & 5.8\% & 0.0\% \\
		
		DarkNet53~\cite{behley2019iccv}& 16.1 & 0.063s & 378B & 50M & 82.0\% & 17.2\% & 15.2\% & 0.8\% & 6.1\% & 68.5\% & 0.0\% & 15.5\% & 63.8\% & 0.4\% & 0.3\% & 0.0\% & 17.3\% & 23.8\% & 13.3\% & 35.6\% & 0.0\% & 6.3\% & 0.0\% \\
		
		%+ RGB feature & 15.0 & 0.110s & 378B & 50M & 19.4\% & 18.1\% & 0.9\% & 7.8\% & 55.5\% & 1.8\% & 17.9\% & 61.0\% & 0.8\% & 10.6\% & 0.0\% & 21.3\% & 37.1\% & 12.5\% & 40.9\% & 0.0\% & 14.8\% & 0.0\% \\
		
		\hline
		
		Unet w/ Cartesian BEV& 49.5 & 0.028s & 60B & 14M &  83.5\% & 20.3\% & 27.0\% & 7.3\% & 20.3\% & 66.0\% & 1.9\% & 25.2\% & 54.7\% & 6.5\% & 12.7\% & 0.0\% & 20.3\% & 26.8\% & 21.4\% & 42.5\% & 0.0\% & 9.5\% & 0.0\% \\
		
		%+ RGB feature & 32.8 & 0.032s & 60B & 14M & \textbf{27.7\%} & 29.4\% & 9.7\% & 27.2\% & 81.7\% & 0.2\% & 54.5\% & 66.3\% & 1.3\% & 17.4\% & 0.0\% & 21.7\% & 46.1\% & 26.7\% & 50.7\% & 0.0\% & 17.5\% & 0.0\% \\
		
		PolarNet& 38.4 & 0.031s & 60B & 14M &  \textbf{85.4\%} & 23.9\% & 23.8\% & 10.1\% & 18.2\% & 69.7\% & 9.6\% & 49.1\% & 58.5\% & 0.0\% & 11.3\% & 0.0\% & 28.3\% & 37.6\% & 24.8\% & 42.8\% & 0.0\% & 14.8\% & 0.0\% \\
		
		%+ RGB feature & 28.4 & 0.038s & 60B & 14M & \textbf{27.7\%} & 30.1\% & 11.8\% & 24.5\% & 76.6\% & 3.0\% & 55.8\% & 63.1\% & 6.0\% & 16.5\% & 0.0\% & 26.9\% & 50.4\% & 30.1\% & 51.7\% & 0.0\% & 10.7\% & 0.0\% \\
		
		\hline
		
	\end{tabular}
	}
	\resizebox{\linewidth}{!}{%
	\begin{tabular}{|l|c|*{21}{c}|}
		\hline
		\multirow{2}*{{Model}} & \multirow{2}*{{mIoU}} & \multicolumn{21}{c|}{Per class IoU}  \\ 
		\cline{3-23}
		&& \rotatebox{90}{\makecell{obstacles / \\ trash}} &	\rotatebox{90}{poles} &	\rotatebox{90}{\makecell{RD \\ restricted \\ area}} &	\rotatebox{90}{animals} &	\rotatebox{90}{\makecell{grid \\ structure}} &	\rotatebox{90}{\makecell{signal \\ corpus}} &	\rotatebox{90}{\makecell{drivable \\ cobblestone}} &	\rotatebox{90}{\makecell{electronic \\ traffic}} &	\rotatebox{90}{\makecell{slow drive \\ area}} &	\rotatebox{90}{\makecell{nature \\ object}} &	\rotatebox{90}{\makecell{parking \\ area}} &	\rotatebox{90}{sidewalk} &	\rotatebox{90}{ego car} &	\rotatebox{90}{\makecell{painted \\ driv. instr.}} &	\rotatebox{90}{\makecell{traffic \\ guide obj.}} &	\rotatebox{90}{\makecell{dashed \\ line}} &	\rotatebox{90}{\makecell{RD normal \\ street}} &	\rotatebox{90}{sky} &	\rotatebox{90}{buildings} &	\rotatebox{90}{\makecell{blurred \\ area}} &	\rotatebox{90}{\makecell{rain dirt}}\\
		\hline 
		\hline
		
		Squeezeseg~\cite{wu2018squeezeseg}& 8.9\% & 0.0\% & 0.3\% & 0.0\% & 0.0\% & 0.0\% & 0.0\% & 0.0\% & 0.0\% & 0.0\% & 64.5\% & 0.0\% & 13.7\% & 0.0\% & 0.0\% & 0.1\% & 0.2\% & 77.7\% & 10.4\% & 27.7\% & 0.0\% & 0.0\%\\

		Squeezesegv2~\cite{wu2019squeezesegv2}& 16.4\% & 0.2\% & 5.2\% & 29.5\% & 0.0\% & 10.3\% & 5.5\% & 2.7\% & 0.0\% & 1.9\% & 76.4\% & 3.8\% & 29.2\% & 0.0\% & 6.4\% & 12.4\% & 17.1\% & 85.8\% & 12.1\% & 50.9\% & 0.0\% & 0.0\% \\
		
		DarkNet53~\cite{behley2019iccv}& 17.2\% & 3.9\% & 7.6\% & 38.7\% & 0.0\% & 10.8\% & 4.4\% & 3.3\% & 0.0\% & 0.0\% & 77.9\% & 3.1\% & 31.5\% & 0.0\% & 9.4\% & 7.3\% & 15.7\% & 86.4\% & 12.9\% & 55.2\% & 0.0\% & 0.0\%\\
		
		%+ RGB feature & 19.4\% & 5.6\% & 8.7\% & 39.9\% & 0.0\% & 12.1\% & 10.2\% & 5.1\% & 0.0\% & 1.0\% & 80.9\% & 4.8\% & 41.4\% & 0.0\% & 11.0\% & 9.0\% & 24.6\% & 87.5\% & 25.4\% & 53.8\% & 0.0\% & 0.0\%\\
		
		\hline
		
		Unet w/ Cartesian BEV& 20.3\% & 4.3\% & 11.0\% & 44.7\% & 0.0\% & 11.8\% & 11.9\% & 6.4\% & 0.0\% & 0.0\% & 81.6\% & 11.9\% & 35.1\% & 0.0\% & 6.9\% & 13.7\% & 20.2\% & 89.2\% & 5.8\% & 56.1\% & 0.0\% & 0.0\% \\
		
		%+ RGB feature & \textbf{27.7\%} & 10.4\% & 12.2\% & 46.3\% & 0.0\% & 18.8\% & 15.9\% & 10.7\% & 0.0\% & 13.6\% & 85.9\% & 16.1\% & 47.7\% & 0.0\% & 17.2\% & 21.1\% & 30.9\% & 90.4\% & 16.1\% & 65.4\% & 0.0\% & 0.0\%\\
		
		PolarNet& 23.9\% & 8.0\% & 11.0\% & 55.6\% & 0.0\% & 14.8\% & 11.9\% & 7.0\% & 0.0\% & 4.4\% & 81.6\% & 12.8\% & 42.5\% & 0.0\% & 12.7\% & 11.5\% & 31.8\% & 90.3\% & 9.2\% & 57.0\% & 0.0\% & 0.0\% \\
		
		%+ RGB feature & \textbf{27.7\%} & 11.5\% & 13.4\% & 50.2\% & 0.0\% & 19.1\% & 16.5\% & 11.3\% & 0.0\% & 2.0\% & 84.9\% & 12.1\% & 52.9\% & 0.0\% & 19.5\% & 15.4\% & 36.7\% & 91.3\% & 13.3\% & 62.4\% & 0.0\% & 0.0\%\\
		
		\hline
	\end{tabular}
	}
\vspace{-3pt}
\end{table*}

\begin{table*}
	\centering
	\caption{How projection methods impact models' segmentation performance on \textbf{val} split of SemanticKITTI.}  
	\label{tab:SKITTI_val}
	\resizebox{\linewidth}{!}{
	\begin{tabular}{|l|l|c|c|c|c|c|*{19}{c}|}
		\hline
		\multirow{2}*{Model} & \multirow{2}*{Projection} & \multirow{2}*{FPS} & \multirow{2}*{Latency} & \multirow{2}*{MACs} & \multirow{2}*{Params} & \multirow{2}*{mIoU} & \multicolumn{19}{c|}{Per class IoU}  \\ 
		\cline{8-26}
		&&&&&&&\rotatebox{90}{car} &	\rotatebox{90}{bicycle} &	\rotatebox{90}{motorcycle} &	\rotatebox{90}{truck}&	\rotatebox{90}{other-vehicle} &	\rotatebox{90}{person} &	\rotatebox{90}{bicyclist} &	\rotatebox{90}{motorcyclist} &	\rotatebox{90}{road} &	\rotatebox{90}{parking} & \rotatebox{90}{sidewalk} &	\rotatebox{90}{other-ground} &	\rotatebox{90}{building} &	\rotatebox{90}{fence} &	\rotatebox{90}{vegetation} &	\rotatebox{90}{trunk} &	\rotatebox{90}{terrain} &	\rotatebox{90}{pole} &	\rotatebox{90}{traffic-sign} \\
		\hline 
		\hline
		
		\multirow{3}*{Squeezeseg}& Spherical& 83.6 & 0.012s & 14B & 0.9M & 31.8\% & 79.4\% & 0.0\% & 0.0\% & 3.2\% & 1.3\% & 0.0\% & 0.0\% & 0.0\% & 90.9\% & 19.8\% & 74.7\% & 0.0\% & 75.3\% & 31.6\% & 80.6\% & 37.3\% & 71.1\% & 13.2\% & 26.3\% \\

		& Cartesian  BEV& 19.5 & 0.051s & 101B & 1.5M & \textbf{42.6\%} &90.4\% & 15.2\% & 16.6\% & 13.5\% & 16.8\% & 39.0\% & 45.8\% & 0.0\% & 85.7\% & 25.3\% & 65.2\% & 0.0\% & 86.1\% & 32.1\% & 79.7\% & 54.4\% & 60.1\% & 50.9\% & 33.2\% \\

		\rowcolor{gray!40}\cellcolor{white} & Polar BEV& 17.8 & 0.056s & 105B & 1.5M & 42.2\% & 89.8\% & 22.1\% & 19.8\% & 14.2\% & 9.2\% & 37.0\% & 14.3\% & 0.4\% & 83.7\% & 15.8\% & 65.6\% & 0.0\% & 85.9\% & 40.2\% & 85.6\% & 54.2\% & 72.1\% & 54.9\% & 36.7\% \\
		\hline
		
		\multirow{3}*{Resnet-FCN}& Spherical& 38.6 & 0.048s & 92B & 117M & 41.6\% & 82.3\% & 1.5\% & 13.7\% & 65.8\% & 15.5\% & 20.3\% & 31.2\% & 0.0\% & 92.1\% & 32.4\% & 75.6.2\% & 0.1\% & 77.3\% & 31.6\% & 78.1\% & 43.9\% & 66.8\% & 36.6\% & 25.2\% \\

		& Cartesian BEV& 11.7 & 0.088s & 197B & 117M & 49.2\% & 89.9\% & 28.2\% & 15.6\% & 56.5\% & 30.5\% & 41.0\% & 66.1\% & 0.0\% & 88.6\% & 38.3\% & 71.5\% & 6.1\% & 86.5\% & 30.4\% & 81.5\% & 52.2\% & 65.7\% & 46.7\% & 39.3\% \\

		\rowcolor{gray!40}\cellcolor{white} & Polar BEV& 11.5 & 0.091s & 200B & 117M & \textbf{52.5\%} & 92.1\% & 22.8\% & 36.2\% & 57.5\% & 24.6\% & 42.5\% & 63.9\% & 0.0\% & 92.1\% & 43.6\% & 77.5\% & 1.7\% & 90.0\% & 46.9\% & 84.4\% & 56.0\% & 73.1\% & 53.3\% & 40.2\% \\
		\hline
		
		\multirow{3}*{DRN-DL}& Spherical& 39.1 & 0.038s & 94B & 41M & 43.4\% & 82.6\% & 3.1\% & 24.5\% & 51.1\% & 18.3\% & 27.3\% & 23.9\% & 0.0\% & 93.0\% & 37.2\% & 77.4\% & 0.2\% & 76.8\% & 42.1\% & 79.7\% & 46.2\% & 68.7\% & 39.2\% & 32.9\% \\

		& Cartesian BEV& 10.0 & 0.100s & 171B & 41M & 46.7\% & 90.4\% & 14.1\% & 20.3\% & 51.4\% & 37.3\% & 39.3\% & 42.3\% & 0.0\% & 87.6\% & 30.6\% & 68.0\% & 1.5\% & 86.5\% & 33.0\% & 83.2\% & 49.2\% & 69.8\% & 44.3\% & 39.0\% \\

		\rowcolor{gray!40}\cellcolor{white} & Polar BEV& 9.9 & 0.101s & 173B & 41M & \textbf{51.2\%} & 91.6\% & 19.4\% & 35.0\% & 34.6\% & 20.8\% & 50.8\% & 55.1\% & 0.0\% & 92.5\% & 38.6\% & 77.5\% & 1.1\% & 88.5\% & 44.4\% & 84.8\% & 59.7\% & 70.6\% & 56.7\% & 40.2\% \\
		\hline
		
		\multirow{3}*{Resnet-DL}& Spherical& 89.5 & 0.031s & 45B & 59M & 41.6\% & 81.0\% & 0.6\% & 17.1\% & 58.9\% & 12.1\% & 21.3\% & 24.7\% & 0.0\% & 92.5\% & 33.5\% & 76.4\% & 0.0\% & 76.0\% & 40.4\% & 78.6\% & 45.7\% & 68.3\% & 35.1\% & 28.6\% \\

		& Cartesian BEV& 11.8 & 0.090s & 107B & 60M & 50.4\% & 92.6\% & 17.8\% & 41.9\% & 62.0\% & 24.2\% & 42.0\% & 66.3\% & 0.0\% & 87.1\% & 27.2\% & 69.6\% & 0.4\% & 87.4\% & 41.5\% & 84.7\% & 54.8\% & 71.0\% & 48.7\% & 39.1\% \\

		\rowcolor{gray!40}\cellcolor{white} & Polar BEV& 11.7 & 0.094s & 109B & 60M & \textbf{53.6\%} & 91.5\% & 30.7\% & 38.8\% & 46.4\% & 24.0\% & 54.1\% & 62.2\% & 0.0\% & 92.4\% & 47.1\% & 78.0\% & 1.8\% & 89.1\% & 45.5\% & 85.4\% & 59.6\% & 72.3\% & 58.1\% & 42.2\% \\
		
		\hline
		
		%\multirow{2}*{Unet}& Conventional BEV& 54.3\% & 93.3\% & 30.3\% & 52.9\% & 44.9\% & 28.0\% & 52.0\% & 79.5\% & 0.0\% & 88.8\% & 35.0\% & 72.2\% & 1.0\% & 89.3\% & 43.9\% & 85.5\% & 61.9\% & 74.5\% & 57.0\% & 42.5\% \\

		%& Polar BEV& \textbf{54.9\%} & 93.3\% & 37.6\% & 39.4\% & 42.0\% & 33.7\% & 51.1\% & 63.5\% & 0.0\% & 93.2\% & 40.5\% & 79.2\% & 1.5\% & 90.4\% & 54.4\% & 85.7\% & 61.6\% & 71.1\% & 61.7\% & 44.1\% \\

	\end{tabular}
	}
	\vspace{-3pt}
\end{table*}

We present our A2D2 results in Table.~\ref{tab:A2D2_test}. Our method undoubtedly outperforms other baselines in terms of both mIoU and speed. By observing mIoU, we see A2D2 to be a challenging dataset. Despite being the leading method, our mIoU using only \lidar data on this dataset is merely 23\% while our mIoU on SemanticKITTI is 54\%. Our methods also double the IoU in multiple classes such as bicycle, pedestrian, small-vehicle, traffic-light, sidebars, signal corpus. parking area and dash-line. The dataset is indeed challenging since both baselines and our methods achieved near zero IoU in multiple classes as well.

\subsection{Paris-Lille-3D Segmentation Experiment}

As indicated by the Paris-Lille-3D segmentation results in Table~\ref{tab:PL3D_test}, PolarNet outperforms DarkNet53 by 3.7\% in mIoU. The segmentation performances are interestingly diverse. PolarNet greatly improved the results in barrier since it is mostly far away from vehicle. However Cartesian Unet has great advantage in the trash can, which has very few samples in both training and validation.

\begin{table}
	\centering
	\caption{Segmentation results on \textbf{test} split of Paris-Lille-3D.}
	\vspace{0pt}
	\label{tab:PL3D_test}
	\resizebox{\linewidth}{!}{%
	\begin{tabular}{|l|c|c|*{9}{c}|}
		\hline
		\multirow{2}*{Model} & \multirow{2}*{Acc} & \multirow{2}*{mIoU} & \multicolumn{9}{c|}{Per class IoU}  \\ 
		\cline{4-12}
		&&&\rotatebox{90}{ground} &	\rotatebox{90}{building} &	\rotatebox{90}{pole} &	\rotatebox{90}{bollard}&	\rotatebox{90}{trash can} &	\rotatebox{90}{barrier} &	\rotatebox{90}{pedestrian} &	\rotatebox{90}{car} &	\rotatebox{90}{vegetation} \\
		\hline 
		\hline
		
		Squeezesegv2~\cite{wu2019squeezesegv2} & 87.3\%& 36.9\%&	95.9\%&	82.7\%&	18.7\%&	9.9\%&	3.8\%&	15.2\%&	3.4\%&	49.9\%&	52.8\%\\
        DarkNet53~\cite{behley2019iccv} & \textbf{88.9}\%& 40.0\%&	96.7\%&	\textbf{84.9\%}&	19.5\%&	16.7\%&	4.8\%&	17.6\%&	3.4\%&	58.2\%&	\textbf{57.9\%}\\
		\hline
		
		Unet w/ Cartesian BEV&80.9\%& 40.3\%&	96.0\%&	44.0\%&	\textbf{38.4\%}&	\textbf{42.8\%}&	\textbf{12.7\%}&	12.4\%&	\textbf{12.1}\%&	70.4\%&	33.60\%\\
		
		PolarNet& 87.5\% & \textbf{43.7\%}  & \textbf{96.8\%} &	69.1\% &	32.2\% &	27.6\% &	2.4\% &	\textbf{27.5\%} &	\textbf{12.1\%}& 	\textbf{74.0\%}	& 51.60\%\\
		
		\hline
	\end{tabular}
	}
\vspace{-3pt}
\end{table}

\subsection{Impact of Projection Methods}

 In Table \ref{tab:SKITTI_val}, we show the results of SemanticKITTI mIoU with different segmentation backbone networks, include SqueezeSeg, Resnet-50-FCN, DRN-DeepLab and Resnet-101-DeepLab, on three different projection methods: spherical projection proposed in SqueezeSeg~\cite{wu2018squeezeseg}, Cartesian BEV and our polar BEV. For spherical projection, we followed the setup of projecting point clouds with zenith angles ranging from $-25^{\circ}$ to $3^{\circ}$ into $[64,2048]$ grids in the projected sphere plane as in \cite{milioto2019rangenet++}. The results show that no matter what segmentation network is used, BEV always considerably outperforms spherical projection methods. The inferior performance of spherical projection can be explained in two ways. First, since point clouds are directly projected onto 2D sphere coordinate, spherical projection suffers more from the error generated from quantization. Second, distance information is lost during projection even when explicitly encoded into features, which enables points distant in space to locate in neighboring 2D grids and easily get misclassified as the same label. Meanwhile, experiments also show that polar BEV achieves a comparable and better performance than Cartesian BEV for each backbone network. Since \lidar point clouds are sparse in space and discontinuous due to occlusion, quantization creates irregular and inconsistent edges in 2D representations. Such inconsistency allows Unet to stand out from those backbone segmentation networks and achieve the best performance.

\subsection{Augmenting \lidar Segmentation}

\begin{table}
\centering
\caption{Improvement break down. \textbf{RC} denotes ring convolution. \textbf{9F} denotes using 9 features to describe each point. \textbf{FA} denotes flip augmentation. \textbf{FS} denotes fixed volume space. \textbf{TG} denotes tuned grid size.}
\label{tab:ablation_study}
\begin{tabular}{ccccc|c}
\hline
\textbf{RC} & \textbf{9F} & \textbf{FA} & \textbf{FS} & \textbf{TG} & mIoU            \\

\hline
   &    &    &    &    &  46.9\%          \\
$\times$  &    &    &    &    &  47.4\%          \\
$\times$  & $\times$  &    &    &    &  48.5\%          \\
$\times$ & $\times$  & $\times$  &    &    &  50.6\%          \\
$\times$  &$\times$  & $\times$  & $\times$  &    &  53.4\%          \\
$\times$ & $\times$  & $\times$  & $\times$  & $\times$  &  \textbf{54.9\%} \\
\hline
\end{tabular}

\vspace{-4pt}
\end{table}

In addition, we analyze the effects of different training settings on the validation mIoU result in Table \ref{tab:ablation_study}. The baseline is our polar BEV Unet network with grid size of $[256,256,32]$. ``RC'' denotes using the ring convolution kernel rather than a normal 2D convolution in the backbone network. ``9F'' denotes we use 2 Cartesian coordinates, 3 residual distances from the center of the assigned grid and 1 reflection in addition to 3 polar coordinates, totaling 9 features as the input of our CNN network for each point. ``FA'' denotes we add 25\% probability each to randomly flip a point cloud along $x$, $y$ and $x+y$ axes for data augmentation. ``FS'' denotes we fix the volume space of BEV based on our statistical analysis mentioned before. ``TG'' denotes we tuned the grid size to be $[480,360,32]$ after trying different grid size configurations to reach the best performance. From Table \ref{tab:ablation_study}, we can see that fixing volume space contributes the most significant improvement of 2.8\% increase in mIoU by making scale invariant in each scan. These augmentations are applied to the Cartesian BEV network as well in all other experiments.

%\textbf{Point number vs. distance}: We count the distribution of points and points per grid in validation split w.r.t the distance in Fig.\ref{}. It displays that point number decreases exponentially with distance, yet polar BEV spreads point cloud onto BEV representation in a more balanced manner.

\subsection{mIoU vs. Distance to Sensor}

\begin{figure}
  \centering
  \includegraphics[width=0.7\linewidth]{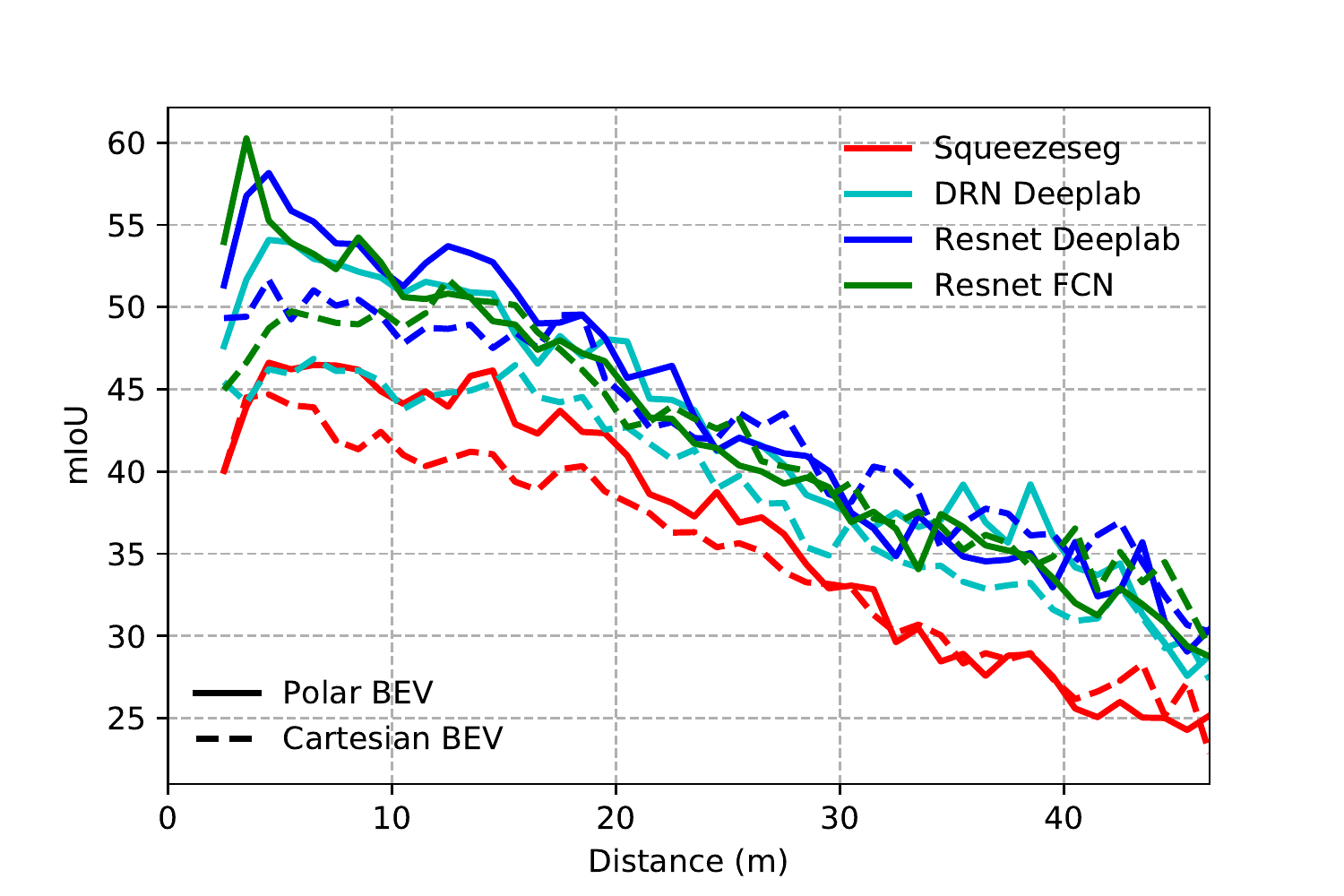}
  \vspace{-4pt}
  \caption{Points distance to sensor vs. their IoU in different networks and projections. Clearly, closer points benefits the most from polar BEV regardless of backbone networks.}
  \label{fig:iou_vs_dist}
  \vspace{-15pt}
\end{figure}

Furthermore, we sort the point-wise predictions in validation split w.r.t. the distance from the sensor and analyze the mIoU result at different distances. Fig.~\ref{fig:iou_vs_dist} shows that with the increase of distance, mIoU reduces simultaneously. The reason for this pattern is that distant points are more rare and separated in space, which makes it harder for the segmentation network to extract contextual information from the BEV representation. This observation is the same as in~\cite{behley2019iccv}. However, the most intriguing conclusion we obtain from this figure relates to the different BEV representations: polar BEV overall gets higher mIoU in close range than Cartesian BEV due to the more evenly distributed points in this BEV representation, as shown in Fig.~\ref{fig:pt_vs_dist}. This grants polar BEV superior mIoU on closer points, which are the majority in a scan.

\iffalse

\subsection{Qualitative Results}

At last, we present our qualitative results in Fig.~\ref{fig:qualitative}. We present both Cartesian BEV segmentation results and polar BEV segmentation results in the first two columns. We also highlight the difference between these two predictions in the third column while the groundtruth labels are shown in the last column. Both BEV methods appear to have very good prediction accuracy at the first glimpse even though their mIoU are different. This indicates the main performance gap between the two is in the details.  The traditional BEV mistakes terrain for sidewalk in row 5. Polar BEV also impressively but falsely predicts a perfectly sharp sidewalk in row 4. Both BEV methods have trouble distinguishing sidewalk and road in most cases.

\begin{figure*}
\centering
  \includegraphics[width=0.9\textwidth]{fig/prediction_visualization.png}
  \caption{Prediction results made by our traditional BEV and Polar BEV networks on SemanticKITTI \textbf{val} split. Best viewed in color.}
  \label{fig:qualitative}
\end{figure*}
\fi